\documentclass[conference]{IEEEtran}
\usepackage{graphicx}
\usepackage{amsmath}
\usepackage{algorithm}
\usepackage{algorithmic}
\usepackage{float}

\title{Optimized Monte Carlo Tree Search for Enhanced Decision Making in the FrozenLake Environment}

\author{\IEEEauthorblockN{Esteban Aldana}
\IEEEauthorblockA{
Computer Science Department, Universidad del Valle de Guatemala \\
Email: ald20591@uvg.edu.gt}
}

\begin{document}
\maketitle

\begin{abstract}
Monte Carlo Tree Search (MCTS) is a powerful algorithm for solving complex decision-making problems. This paper presents an optimized MCTS implementation applied to the FrozenLake environment, a classic reinforcement learning task characterized by stochastic transitions. The optimization leverages cumulative reward and visit count tables along with the Upper Confidence Bound for Trees (UCT) formula, resulting in efficient learning in a slippery grid world. We benchmark our implementation against other decision-making algorithms, including MCTS with Policy and Q-Learning, and perform a detailed comparison of their performance. The results demonstrate that our optimized approach effectively maximizes rewards and success rates while minimizing convergence time, outperforming baseline methods, especially in environments with inherent randomness.
\end{abstract}

\section{Introduction}
Monte Carlo Tree Search (MCTS) is a heuristic search algorithm used extensively in decision-making processes, particularly in domains like game playing, robotics, and optimization problems \cite{browne2012survey}. Its strength lies in its ability to balance exploration and exploitation through randomized sampling, constructing a search tree that guides optimal action selection.

This paper focuses on optimizing MCTS for the \textit{FrozenLake} environment, a standard benchmark in reinforcement learning characterized by its stochastic and slippery dynamics. In FrozenLake, an agent must navigate a grid to reach a goal while avoiding hidden pitfalls, requiring intelligent decision-making under uncertainty.

The primary goal of this study is to enhance the efficiency and effectiveness of MCTS in stochastic environments by integrating cumulative reward and visit count tables, denoted as $Q$ and $N$, respectively. These tables facilitate faster convergence by retaining valuable information from previous explorations. Additionally, we employ the Upper Confidence Bound for Trees (UCT) formula to maintain a strategic balance between exploring new actions and exploiting known rewarding actions.

To validate our approach, we benchmark the optimized MCTS against two other algorithms: MCTS with Policy and Q-Learning. This comparative analysis highlights the advantages of our optimized approach in terms of learning efficiency, performance stability, and execution time.

\section{Related Work}
Monte Carlo Tree Search has been extensively studied and applied across various domains. Browne \textit{et al.} \cite{browne2012survey} provide a comprehensive survey of MCTS methods, highlighting its applications in game playing, robotics, and decision-making under uncertainty. The foundational UCT algorithm introduced by Kocsis and Szepesvári \cite{kocsis2006bandit} introduced a mechanism to balance exploration and exploitation, which has been pivotal in MCTS advancements.

Reinforcement Learning (RL) algorithms, such as Q-Learning \cite{sutton2018reinforcement}, have also been widely used for decision-making tasks. Unlike MCTS, which builds a search tree based on simulations, Q-Learning focuses on learning a value function to guide action selection. While Q-Learning is effective in deterministic environments, its performance can degrade in highly stochastic settings like FrozenLake.

The FrozenLake environment, part of OpenAI Gym \cite{brockman2016openai}, serves as a benchmark for evaluating RL algorithms in stochastic settings. Previous studies have applied both MCTS and Q-Learning to FrozenLake, demonstrating the strengths and limitations of each approach. However, there remains a gap in optimizing MCTS specifically for such environments to enhance its performance and reliability.

\section{Methodology}
The optimized MCTS algorithm introduced in this study aims to improve decision-making in the FrozenLake environment by addressing the inherent challenges of stochasticity and the exploration-exploitation balance. The key innovations include the integration of cumulative reward ($Q$) and visit count ($N$) tables and the application of the UCT formula tailored for stochastic environments.

\subsection{Optimized MCTS Framework}
The optimized MCTS operates through iterative simulations, building a search tree that represents possible state-action trajectories. The use of $Q$ and $N$ tables allows the algorithm to retain and update information about the cumulative rewards and the number of times each action has been explored in a given state. This memory mechanism enhances the algorithm's ability to make informed decisions based on historical performance data.

The UCT formula is central to balancing exploration and exploitation. It is defined as:

\begin{equation}
\text{UCT}(s,a) = \frac{Q(s,a)}{N(s,a)} + c \sqrt{\frac{\ln N(s)}{N(s,a)}}
\end{equation}

where:
\begin{itemize}
    \item $Q(s,a)$ is the cumulative reward for state-action pair $(s,a)$.
    \item $N(s,a)$ is the visit count for state-action pair $(s,a)$.
    \item $N(s)$ is the total visit count for state $s$.
    \item $c$ is the exploration weight parameter.
\end{itemize}

By incorporating the logarithm of the total visit count and the individual action visit counts, the formula dynamically adjusts the exploration term based on the current knowledge of the state-action space. This ensures that actions with higher potential rewards are prioritized while still allowing for the exploration of less-visited actions that may yield better long-term benefits.

\subsection{Pseudocode}

The key steps of the optimized MCTS algorithm are summarized in the following pseudocode:

\begin{algorithm}
\caption{Optimized MCTS Algorithm}
\begin{algorithmic}
\STATE Initialize $Q$ and $N$ tables
\FOR {each episode}
    \STATE Reset environment to initial state
    \STATE Initialize empty path
    \WHILE {not terminal state}
        \STATE Select action using UCT formula
        \STATE Add action to path
        \STATE Step environment to next state
    \ENDWHILE
    \STATE Simulate reward from terminal state
    \STATE Backpropagate reward along the path
\ENDFOR
\end{algorithmic}
\end{algorithm}

This algorithm ensures that the agent systematically explores the action space while progressively favoring actions that have yielded higher rewards in the past.

\subsection{Implementation Details}
The optimized MCTS algorithm was implemented using Python and the OpenAI Gym environment for FrozenLake. The key components of the implementation include:

\begin{itemize}
    \item \textbf{State Representation}: States are represented by the discrete positions on the FrozenLake grid.
    \item \textbf{Action Space}: The action space consists of four possible moves (left, down, right, up).
    \item \textbf{Reward Structure}: The agent receives a reward of 1 upon reaching the goal state and 0 otherwise.
    \item \textbf{Simulation Environment}: A separate simulation environment is used to perform rollouts during the tree policy phase.
    \item \textbf{Hyperparameters}: The exploration weight $c$ is set to 1.4, and the number of simulations per move is 100.
\end{itemize}

\subsection{Comparison with MCTS with Policy and Q-Learning}
To contextualize the effectiveness of the optimized MCTS, we compare it against:

\begin{itemize}
    \item \textbf{MCTS with Policy}: This implementation leverages simulations to update a Q-table that guides policy learning. It balances exploration and exploitation through cumulative reward simulations but is computationally expensive due to numerous simulations.
    \item \textbf{Q-Learning}: A value-based RL algorithm that updates a Q-table based on the Bellman equation. It employs an $\epsilon$-greedy policy for action selection.
\end{itemize}

By benchmarking against these approaches, we aim to demonstrate how the optimized MCTS not only addresses the shortcomings of MCTS with Policy but also offers superior performance in stochastic settings compared to traditional RL methods like Q-Learning.

\section{Benchmarking and Comparison}
To assess the effectiveness of the optimized MCTS algorithm, we conducted a comprehensive comparison with MCTS with Policy and Q-Learning. All algorithms were evaluated in the \textit{FrozenLake} environment over 100,000 episodes, providing a robust dataset for performance analysis.

\subsection{Performance Metrics}
The primary metrics used for comparison include:

\begin{itemize}
    \item \textbf{Success Rate}: The proportion of episodes in which the agent successfully reaches the goal.
    \item \textbf{Average Reward}: The mean reward accumulated per episode.
    \item \textbf{Convergence Rate}: The number of steps taken to reach the goal, indicating the efficiency of the learned policy.
    \item \textbf{Execution Time}: The time taken to complete 100,000 episodes.
\end{itemize}

\subsection{Experimental Results}
The results of the experiments are as follows:

\begin{itemize}
    \item \textbf{Optimized MCTS} achieved an average reward of \textbf{0.8} and a success rate of \textbf{70\%}, stabilizing after approximately \textbf{10,000} episodes. The execution time was \textbf{48.41 seconds}.
    \item \textbf{MCTS with Policy} had a much slower execution time of \textbf{1,758.52 seconds}, with an average reward of \textbf{0.4} and a success rate of \textbf{35\%}. The agent required fewer steps per episode (about \textbf{30 steps}) to converge, but its overall performance in terms of success rate and reward was inferior.
    \item \textbf{Q-Learning} demonstrated slower initial progress but eventually stabilized around an average reward of \textbf{0.8} and a success rate of \textbf{60\%} after \textbf{40,000} episodes. Its execution time was similar to Optimized MCTS at \textbf{42.74 seconds} but required more steps per episode (about \textbf{50 steps}).
\end{itemize}

\subsubsection{Average Reward per Episode}
\begin{figure}[H]
    \centering
    \includegraphics[width=\linewidth]{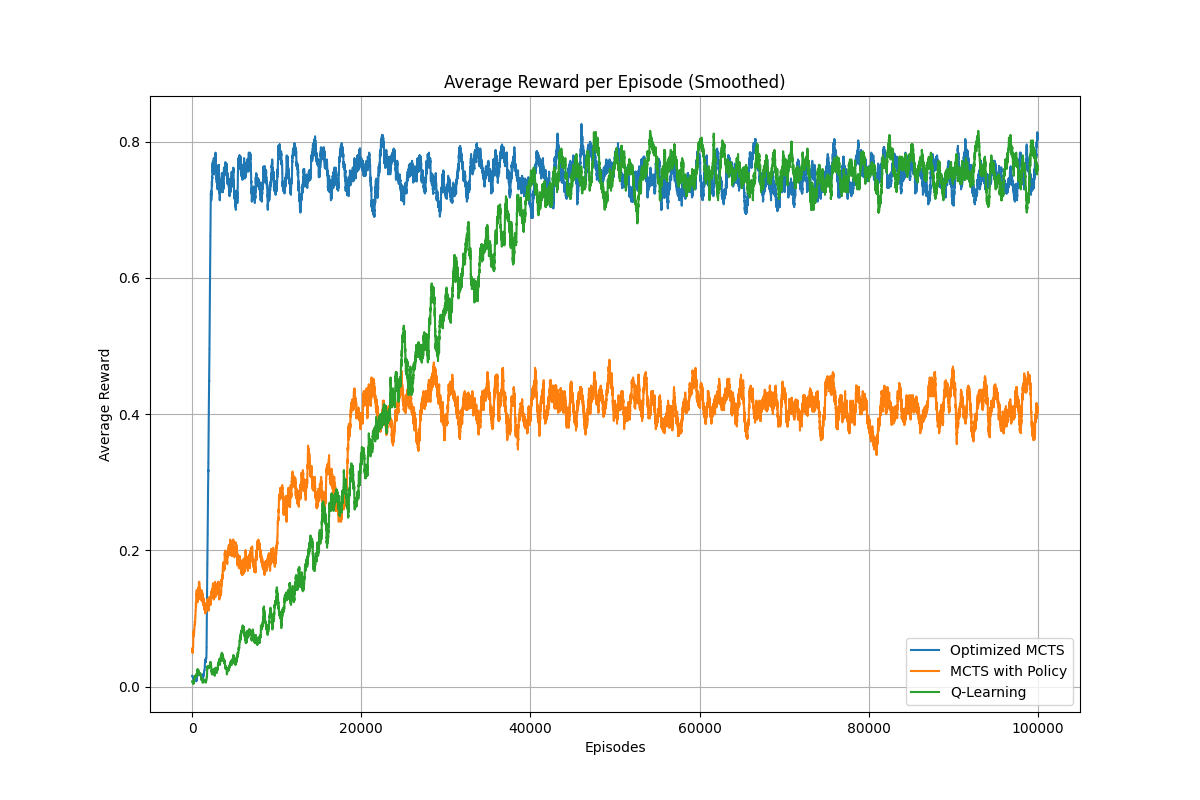}
    \caption{Average Reward per Episode (Smoothed) Comparison}
    \label{fig:average_reward_comparison}
\end{figure}

Figure \ref{fig:average_reward_comparison} shows the smoothed average reward per episode for the three algorithms. The optimized MCTS reaches the highest average reward fastest, followed by Q-Learning, while MCTS with Policy lags behind significantly.

\subsubsection{Convergence Rate}
\begin{figure}[H]
    \centering
    \includegraphics[width=\linewidth]{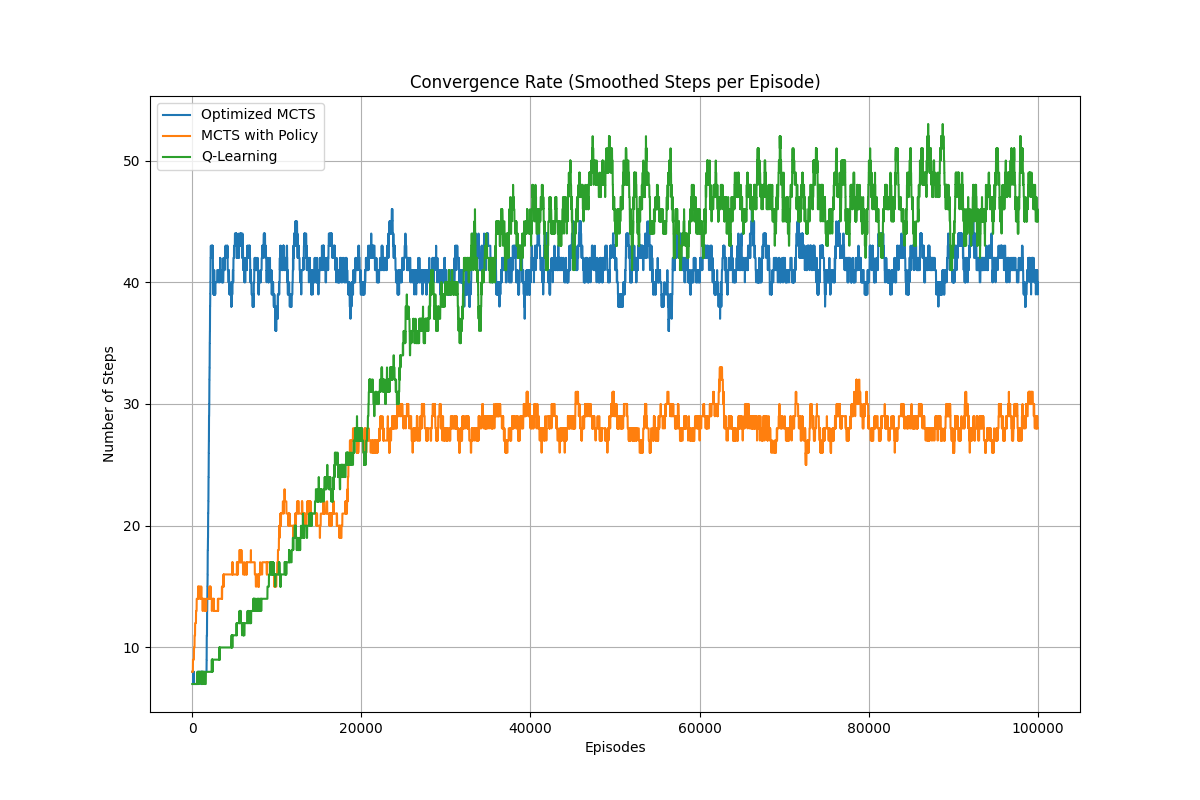}
    \caption{Convergence Rate (Smoothed Steps per Episode) Comparison}
    \label{fig:convergence_rate_comparison}
\end{figure}

Figure \ref{fig:convergence_rate_comparison} presents the convergence rate (steps per episode) for all algorithms. While MCTS with Policy converges with fewer steps, Optimized MCTS and Q-Learning require more steps but yield higher rewards and success rates.

\subsubsection{Success Rate per Episode}
\begin{figure}[H]
    \centering
    \includegraphics[width=\linewidth]{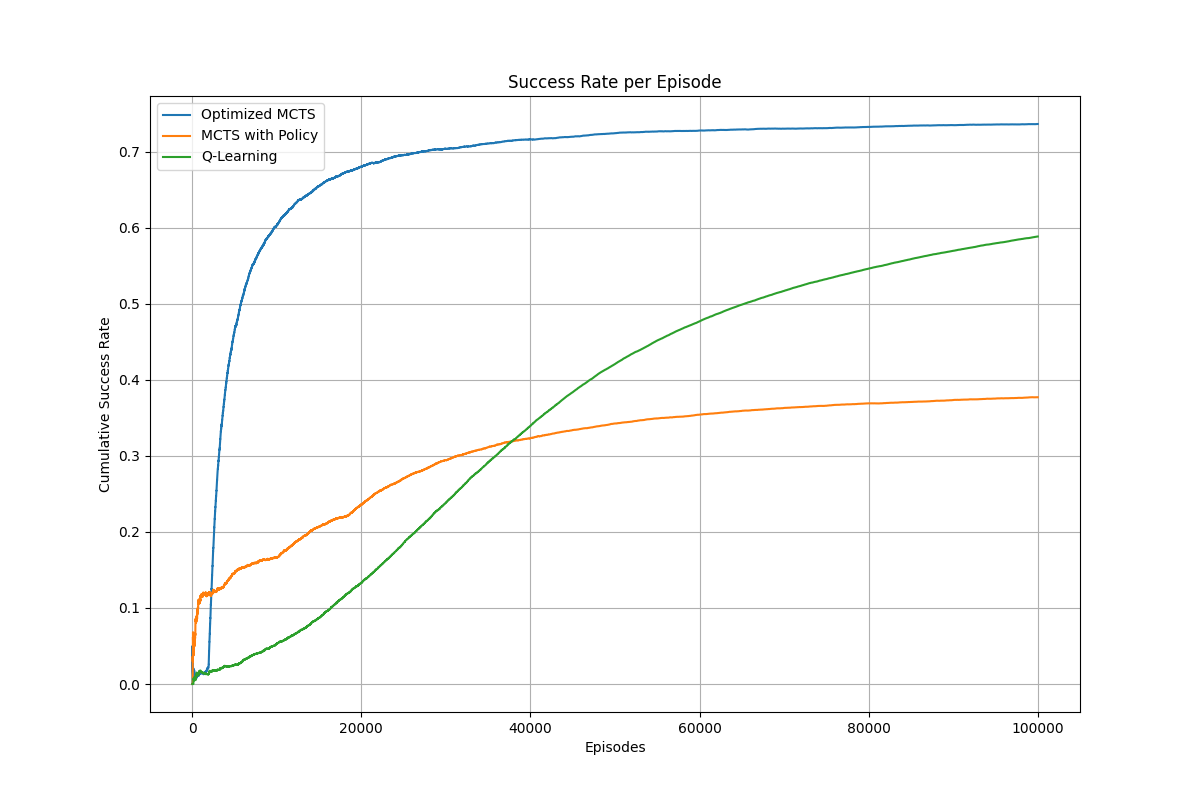}
    \caption{Success Rate per Episode Comparison}
    \label{fig:success_rate_comparison}
\end{figure}

The success rate comparison is shown in Figure \ref{fig:success_rate_comparison}. Optimized MCTS achieves the highest success rate, followed by Q-Learning, while MCTS with Policy has a lower success rate despite its faster convergence in terms of steps per episode.

\subsubsection{Execution Time}
The execution times for each algorithm are summarized in Table \ref{tab:execution_time_comparison}. MCTS with Policy takes significantly longer than both Optimized MCTS and Q-Learning, making it less suitable for scenarios where computational efficiency is a priority.

\begin{table}[H]
\centering
\caption{Execution Time per Algorithm}
\label{tab:execution_time_comparison}
\begin{tabular}{lc}
\hline
\textbf{Algorithm} & \textbf{Execution Time (seconds)} \\
\hline
Optimized MCTS & 48.41 \\
MCTS with Policy & 1,758.52 \\
Q-Learning & 42.74 \\
\hline
\end{tabular}
\end{table}

\section{Conclusion}
This study demonstrates the effectiveness of the optimized Monte Carlo Tree Search (MCTS) algorithm in solving the FrozenLake environment. By utilizing cumulative reward and visit count tables alongside the Upper Confidence Bound for Trees (UCT) formula, the algorithm successfully learns a policy that maximizes the agent's rewards and success rate while minimizing the number of steps required per episode.

The results show that the agent stabilizes around an average reward of 0.8 and a success rate of 70\%, with a consistent convergence rate of 40 steps per episode. These metrics reflect the robustness of the optimized MCTS in managing exploration and exploitation in stochastic environments. Compared to MCTS with Policy and Q-Learning, the optimized approach offers significant improvements in performance and learning efficiency while maintaining competitive execution times.

Future work could explore enhancing the exploration phase through techniques like adaptive exploration constants or integrating model-based strategies to further improve the agent's success rate and reduce performance fluctuations.

\end{document}